\pdfoutput=1

\documentclass[11pt]{article}

\usepackage[]{acl}

\usepackage{times}
\usepackage{latexsym}

\usepackage[T1]{fontenc}

\usepackage[utf8]{inputenc}

\usepackage{microtype}
\usepackage{graphicx}
\usepackage{url}
\usepackage{hyperref}
\usepackage{times}
\usepackage{latexsym}
\usepackage{array}
\usepackage{multirow}
\usepackage{amssymb}
\usepackage{amsmath}
\usepackage{mathpartir}
\usepackage{mathtools}
\usepackage{amsthm}
\usepackage{xspace}
\usepackage{tikz-dependency}
\usepackage{cancel}
\usepackage{wrapfig}
\usepackage{dsfont}
\usepackage{booktabs}
\usepackage{subfiles}
\usepackage{enumitem}
\usepackage{framed}
\usepackage{footnote}
\usepackage{arydshln}
\usepackage[linesnumbered,ruled,algo2e]{algorithm2e}
\usepackage{algorithm}
\usepackage{algorithmic}
\usepackage{booktabs, multicol, multirow}
\usepackage{listings}
\usepackage{nicefrac}
\usepackage{bm}
\usepackage[version=4]{mhchem}
\usepackage{tabularx}
\usepackage{siunitx}

\newcommand{\tuple}[1]{\ensuremath{\langle {#1} \rangle}}

\newcommand{\notes}[1]{}

\theoremstyle{definition}

\theoremstyle{plain}

\newcommand{\ith}[1]{\ensuremath{i^{{th}}}}

\newcount\permx
\newcount\permy
\def\permdot#1#2{
\permx=#1 \advance\permx by-1
\permy=#2 \advance\permy by-1
\psframe[fillcolor=black, fillstyle=solid]
(\permx,\permy)(#1, #2)
}

\newcommand{\boxnum}[1]{{\setlength{\fboxsep}{1pt}\raisebox{1pt}{\hspace{1pt}\fbox{\tiny #1}\hspace{1pt}}}}
\newcommand{\ind}[1]{\ensuremath{_{\kern-0.5pt\boxnum{#1}}}}

\newcommand{\smallnt}[1]{\ensuremath{_{\mbox{\tiny PP}}}\xspace}

\iffalse

\else

\fi

\newcommand{\bos}{\mbox{\scriptsize \texttt{<s>}}\xspace}
\newcommand{\eos}{\mbox{\scriptsize \texttt{</s>}}\xspace}

\newcolumntype{Y}{>{\small\raggedright\arraybackslash}X}
\newcolumntype{J}{>{\small\hsize=2.3\hsize}X}
\newcolumntype{K}{>{\small\hsize=.3\hsize}X}
\newcolumntype{L}{>{\small\hsize=.1\hsize}X}

\title{Language-Informed Beam Search Decoding for Multilingual Machine Translation}

\author{
    Yilin Yang$^{1, }$\thanks{\ \ Work done mostly at Oregon State University.},
    Stefan Lee$^2$,
    Prasad Tadepalli$^2$ \\
    $^1$Meta AI, \ \ \ 
    $^2$Oregon State University \\
    \texttt{yilinyang721@gmail.com} \\
}

\begin{document}
\maketitle
\begin{abstract}
Beam search decoding is the de-facto method for decoding auto-regressive Neural Machine Translation (NMT) models, including multilingual NMT where the target language is specified as an input.
However, decoding multilingual NMT models commonly produces ``off-target'' translations -- yielding translation outputs not in the intended language.
In this paper, we first conduct an error analysis of off-target translations for a strong multilingual NMT model and identify how these decodings are produced during beam search. 
We then propose Language-informed Beam Search (LiBS), a general decoding algorithm incorporating an off-the-shelf Language Identification (LiD) model into beam search decoding to reduce off-target translations. 
LiBS is an inference-time procedure that is NMT-model agnostic and does not require any additional parallel data. 
Results show that our proposed LiBS algorithm on average improves +1.1 BLEU and +0.9 BLEU on WMT and OPUS datasets, and reduces off-target rates from 22.9\% to 7.7\% and 65.8\% to 25.3\% respectively.\footnote{\url{https://github.com/yilinyang7/fairseq_multi_fix}}
\end{abstract}

\section{Motivation}
With Neural Machine Translation (NMT)~\cite{bahdanau2014neural,vaswani2017attention} becoming the state-of-the-art approach in the bilingual Machine Translation literature, Multilingual Neural Machine Translation (MNMT) has attracted much attention~\cite{johnson2017google}.
MNMT has two main advantages: a) it enables one model to translate between multiple language pairs and thus reduces the model and deployment complexity from $O(N^2)$ to $O(1)$, and b) it enables transfer learning between high-resource and low-resource languages.
One attractive feature of such transfer learning is \textit{zero-shot} translation, where the multilingual model is able to translate between language pairs unseen during training. For example, after training from French to English and English to German MT data, the model could directly translate French to German.

Despite the theoretical benefits, recent studies have found an overwhelming amount of \textit{off-target translation} especially for the zero-shot directions~\cite{zhang2020improving,yang2021improving}, where the translation is not in the intended language.
Existing methods all aim to mitigate off-targets during training.
\citet{gu2019improved,zhang2020improving} apply Back Translation (BT) to generate synthetic training data for the zero-shot pairs. \citet{yang2021improving} introduces a language prediction loss and regularizes the training gradients with a held-out \textit{oracle} set.
Yet, none of the previous work has investigated the off-target issue at decoding time, i.e. how \textit{off}-target translations emerge and come to outscore \textit{on}-target translations during beam search decoding.

In this work, we first examine when and how off-target translation emerges during beam search decoding, and
then propose Language-informed Beam Search (LiBS), a general algorithm to reduce off-target generation during beam search decoding by incorporating an off-the-shelf Language Identification (LiD) model.
Our experiment results on two large-scale popular MNMT datasets (i.e. WMT and OPUS) demonstrate the effectiveness of LiBS in both reducing off-target rates and improving general translation performance.  On average LiBS reduces off-target rates from 22.9\% to 7.7\% and 65.8\% to 25.3\% on WMT and OPUS respectively, which translates to +1.1 BLEU and +0.9 BLEU overall quality improvement.
Moreover, LiBS can be added post-hoc to reduce off-target translation of any existing multilingual model without requiring any additional data or training.



\section{Experiment Setup}
In this section, we illustrate the data and model setup we used, and the experimental results of our Language-informed Beam Search algorithm.
Since we re-use the exact same model and data setup as ~\citet{wang2020multi,yang2021improving}, and we will provide a brief summary and direct readers to those works for more details.
\subsection{Dataset}
Following~\cite{wang2020multi,yang2021improving}, we conduct experiments on two widely used large-scale MNMT datasets WMT\footnote{Referred to as ``WMT-10'' in \cite{wang2020multi,yang2021improving}, we denoted it as WMT to disambiguate against the WMT 2010 campaign.} and OPUS-100\footnote{We use the deduplicated version from \cite{yang2021improving}.}, where the WMT dataset is concatenated from previous year WMT training data including English and 10 other languages.
Since the WMT competition does not come with zero-shot evaluation data, we use the human labeled multi-way aligned test set from~\cite{yang2021improving}, based on the WMT-19 test set.

\subsection{Model Training and Evaluation}
For both WMT and OPUS-100, we tokenize the dataset with the SentencePiece model~\cite{kudo2018sentencepiece} to form a shared vocabulary of $64$k tokens.
We adopt the Transformer-Big setting~\cite{vaswani2017attention} in our experiments on the open-sourced Fairseq codebase\footnote{\url{https://github.com/facebookresearch/fairseq}}~\cite{ott2019fairseq}.
The model is optimized using the Adam optimizer~\cite{kingma2014adam} with a learning rate of $5 \times 10^{-4}$, 4000 warm-up steps, and a total of 50k training steps.
The multilingual model is trained on 8 V100 GPUs with a batch size of 8192 tokens and gradient accumulation of 8 steps, which essentially simulates the training on 64 V100 GPUs.
To evaluate the baseline model, we employ beam search decoding with a beam size of 5 and a length penalty of 1.0. The BLEU score is then measured by the de-tokenized case-sensitive SacreBLEU\footnote{BLEU+case.mixed+lang.{src}-{tgt}+numrefs.1+smooth.exp+tok.13a+version.1.4.14}~\cite{post2018call}.
To demonstrate meaningful quality improvement, we also report COMET score~\cite{rei2020comet}\footnote{We used the default ``Unbabel/wmt22-comet-da'' model} on the WMT test set.

To evaluate the off-target rates, we borrow the off-the-shelf LiD model\footnote{https://dl.fbaipublicfiles.com/fasttext/supervised-models/lid.176.bin} from FastText~\cite{joulin2016fasttext} to detect the language for system translations. Similar to~\cite{yang2021improving}, we observe an overwhelming off-target rate (averaging 22.9\%) across zero-shot pairs on our strong baseline model.

\section{Analyzing Off-Target Occurrence During Beam Search}
To understand the off-target occurrence during beam search, we analyze the off-target error types on different language pairs, and conduct experiments with varying beam sizes.

\subsection{Multilingual Beam Search Curse}
The beam search curse phenomenon~\cite{yang2018breaking} is widely observed in bilingual NMT models.
Given a larger beam size, the beam search process would explore a larger search space and choose from a larger candidate pool.
Yet empirically, translation performance usually drops significantly with increasing beam sizes.
In our study, we also found this phenomenon prevailing in the multilingual system and highly related to the off-target translation error.

As an example, we demonstrate the beam search curse on WMT De$\to$Fr and Cs$\to$De translation, since both are between high-resource languages and with decent translation performance (between 15 to 20 BLEU).

\begin{table}[t]
    \centering
    \scalebox{0.9}{
	\begin{tabular}{cccc}
        \toprule
	    Direction & Beam size & BLEU & Off-Target Rate \\
        \midrule
        \multirow{3}{*}{De$\to$Fr} & 5 & 17.3 & 23.1\% \\
        & 10 & 16.1 & 31.7\% \\
        & 20 & 14.3 & 41.4\% \\
        \midrule
        \multirow{3}{*}{Cs$\to$De} & 5 & 15.4 & 12.3\% \\
        & 10 & 15.0 & 17.4\% \\
        & 20 & 14.2 & 22.0\% \\
        \bottomrule
        \end{tabular}
    }
    \caption{Multilingual beam search curse on WMT De$\to$Fr and Cs$\to$De, where larger beam widths consistently lead to more off-target translations.}
    \label{tab:mnmt_curse}
\end{table}

Table~\ref{tab:mnmt_curse} illustrates the results on WMT De$\to$Fr and Cs$\to$De. We could clearly observe that the off-target rate grows sub-linearly with the beam size, and as a result the BLEU score drops significantly with increasing beam sizes.
It then raises the curious question of why the off-target rate increases drastically with larger beam sizes, and whether the performance drop (i.e. BLEU decrease) is mainly due to the off-target errors.

\subsection{Off-Target Error Analysis}
As part of a detailed analysis, we study the off-target error type between six zero-shot pairs (i.e. 12 translation directions) from the WMT dataset.
We categorize the off-target errors into three types: translating into English\footnote{English is never the correct target language in our 12 studied translation directions.}, translating into source language, and others.

\begin{table*}[h]
	\centering
	\begin{tabular}{ccccccccc}
        \toprule
	    \multirow{2}{*}{Directions} & \multicolumn{4}{c}{$b=5$} & \multicolumn{4}{c}{$b=20$} \\
        \cmidrule(lr){2-5}\cmidrule(lr){6-9}
        & Total & $\to$English & $\to$Source & Others & Total & $\to$English & $\to$Source & Others \\
        \midrule
        De$\to$Fr & 23.1\% & 11.8\% & 11.1\% & 0.2\% & 41.4\% & 18.5\% & 22.8\% & 0.1\% \\
        Fr$\to$De & 39.9\% & 10.5\% & 29.4\% & 0.0\% & 62.7\% & 17.2\% & 45.5\% & 0.0\% \\
        Cs$\to$De & 12.3\% & 8.5\% & 3.6\% & 0.2\% & 22.0\% & 17.3\% & 4.5\% & 0.2\% \\
        De$\to$Cs & 19.0\% & 2.5\% & 15.8\% & 0.7\% & 27.6\% & 5.9\% & 21.3\% & 0.4\% \\
        De$\to$Ro & 1.6\% & 0.8\% & 0.5\% & 0.3\% & 1.9\% & 1.1\% & 0.5\% & 0.3\% \\
        Ro$\to$De & 7.3\% & 5.9\% & 0.7\% & 0.7\% & 16.3\% & 14.8\% & 0.7\% & 0.8\% \\
        Fr$\to$Et & 22.5\% & 8.1\% & 12.4\% & 2.0\% & 30.6\% & 13.6\% & 15.6\% & 1.4\% \\
        Et$\to$Fr & 26.1\% & 16.5\% & 6.3\% & 3.3\% & 36.6\% & 26.2\% & 6.7\% & 3.7\% \\
        Ro$\to$Et & 10.8\% & 6.3\% & 1.5\% & 3.0\% & 14.8\% & 10.4\% & 1.6\% & 2.8\% \\
        Et$\to$Ro & 2.0\% & 0.5\% & 0.2\% & 1.3\% & 1.9\% & 0.6\% & 0.3\% & 1.0\% \\
        Tr$\to$Gu & 73.7\% & 73.3\% & 0.2\% & 0.2\% & 78.7\% & 78.1\% & 0.2\% & 0.4\% \\
        Gu$\to$Tr & 36.4\% & 35.6\% & 0.1\% & 0.7\% & 41.4\% & 39.6\% & 0.0\% & 1.8\% \\
        Average & 22.9\% & 15.0\% & 6.8\% & 1.1\% & 31.3\% & 20.3\% & 10.0\% & 1.1\% \\
        \bottomrule
	\end{tabular}
	\caption{Off-Target error analysis on 12 WMT zero-shot directions, where most are either $\to$English or $\to$Source.}
	\label{tab:error_type}
\end{table*}

The detailed off-target error analysis of WMT zero-shot direction is shown in Table~\ref{tab:error_type}.
We find that even though the off-target error is overwhelming across languages, it could easily be categorized into mostly two types: translating into English and ``translating'' into source. The ``Others'' error type only comprises a negligible 1.1\% of cases, given the FastText LiD model has an error margin of 0.81\%~\cite{yang2021improving}.

\begin{table*}[h]
	\centering
	\begin{tabularx}{\textwidth}{X|X}
        \toprule
	    \textbf{Source (Fr)} & \textbf{System Output ($\to$De)}\\
        \midrule
        Sa décision a laissé tout le monde sans voix. & Sa décision a laissé tout le monde sans voix. \\ \midrule
        Abandonnez Chequers et commencez à écouter{\color{red}.} » & Abandonnez Chequers et commencez à écouter » \\ \midrule
        {\color{red}«} C’est une très bonne chose {\color{red}»}, {\color{red}dit} Jaynes. & C’est une très bonne chose, {\color{red}sagt} Jaynes. \\
        \bottomrule
	\end{tabularx}
	\caption{Case studies for ``$\to$Source'' errors. We sample three source-translation pairs from the WMT Fr$\to$De test set (with translation LiD-ed as French). Token differences are colored in red.}
	\label{tab:copy_cases}
\end{table*}

\begin{figure}[t]\centering
    \includegraphics[width=0.53\textwidth]{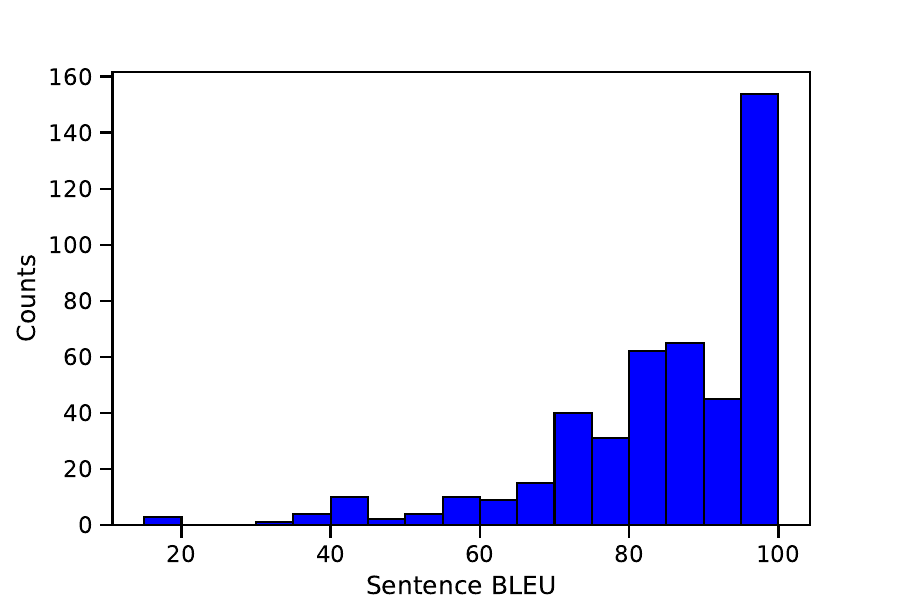}
    \caption{The sentence BLEU distribution between source and system translation from WMT Fr$\to$De ``$\to$Source'' errors, with an average BLEU of 85.3.}
    \label{fig:copy_bleu}
\end{figure}

\paragraph{``$\to$Source'' errors}
We hypothesize that this error is related to the previously studied ``source copying'' behavior~\cite{ott2018analyzing} on the bilingual NMT model.
We then sample three cases from this error type (shown in Table~\ref{tab:copy_cases}).
The case study confirms that the ``$\to$Source'' error type is the same as source copying behavior on bilingual models for these cases.
To quantify the degree of source copying, we run Sentence BLEU evaluation\footnote{We use the \texttt{sentence\_bleu} function from~\cite{post2018call} with \texttt{smooth\_method=`floor'}: \url{https://github.com/mjpost/sacrebleu/blob/master/sacrebleu/compat.py}} between source and system translation on WMT Fr$\to$De ``$\to$Source'' errors.
The sentence BLEU distribution is shown in Figure~\ref{fig:copy_bleu} with an average sentence BLEU of 85.3.
It clearly demonstrates that the ``$\to$Source'' error strongly displays a source copying behavior and is somehow promoted by larger beam sizes.

\paragraph{``$\to$English'' errors}
\begin{table*}[ht]
	\centering
	\begin{tabularx}{\textwidth}{X|X|X}
        \toprule
	    \textbf{Source (Fr)} & \textbf{System Output ($\to$De)} & \textbf{System Output ($\to$En)}\\
        \midrule
        Comme la campagne était très avancée, elle avait pris du retard dans la collecte de fonds, et a donc juré qu’elle ne participerait pas à moins de recueillir 2 millions de dollars. &
        Since the campaign was very advanced, it had {\color{red}fallen} behind in {\color{red}the collection of} funds{\color{red}, }and therefore swore that it would not participate {\color{red}to less than raise} 2 million {\color{red}dollars}. &
        Since the campaign was very advanced, it had {\color{red}lagged} behind in {\color{red}raising} funds and therefore swore that it would not participate {\color{red}unless it raised \$}2 million.
        \\ \midrule
        Woods a perdu ses quatre matchs en France et détient maintenant un record de 13-21-3 en carrière en Ryder Cup. &
        Woods {\color{red}has} lost his four matches in France and now holds a record of 13-21-3 {\color{red}in career} in Ryder Cup. &
        Woods lost his four matches in France and now holds a record of 13-21-3 in {\color{red}the} Ryder Cup.
        \\ \midrule
        Le couple réfute être raciste, et assimile les poursuites à une « extorsion ». &
        The couple refutes being racist{\color{red},} and {\color{red}assimilates} prosecutions {\color{red}to a} ``{\color{red}repression}''. &
        The couple refutes being racist and {\color{red}treats} prosecutions as ``{\color{red}extortion}''. \\
        \bottomrule
	\end{tabularx}
	\caption{Case studies for ``$\to$English'' errors. We sample three source-translation pairs from the WMT Fr$\to$De test set (with translation LiD-ed as English). As a comparison, we also show the output when the model is asked to translate into English. Token differences are colored in red.}
	\label{tab:toen_cases}
\end{table*}

\begin{figure}[ht]\centering
    \includegraphics[width=0.53\textwidth]{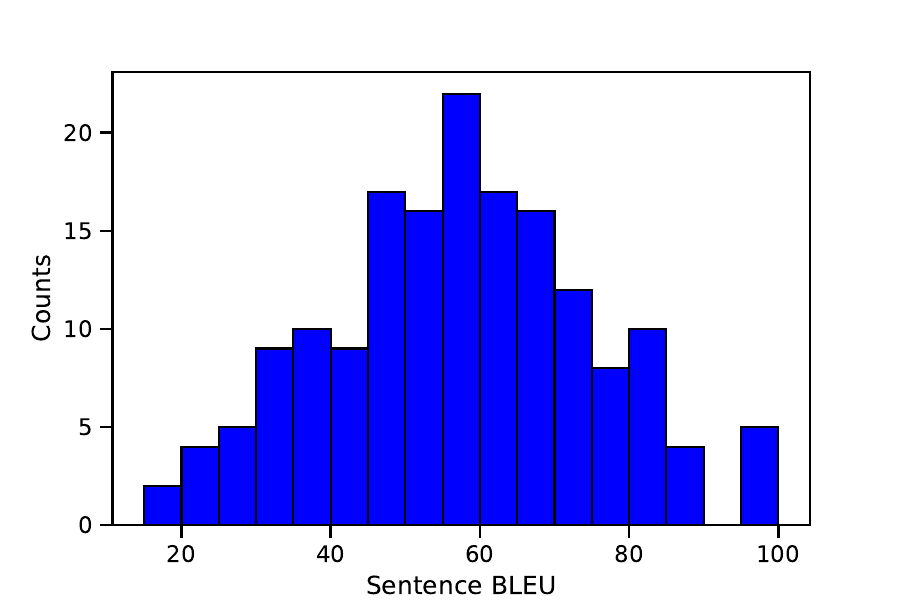}
    \caption{The sentence BLEU distribution between WMT Fr$\to$De ``$\to$English'' errors and Fr$\to$En translation with the same source. The average BLEU is 55.9.}
    \label{fig:toen_bleu}
\end{figure}

\begin{table}[t]
	\centering
	\begin{tabular}{cccc}
        \toprule
	    Direction & BLEU & chrF2 & TER* \\
        \midrule
        Fr$\to$De & 26.92 & 0.572 & 0.62 \\
        Fr$\to$En & 34.91 & 0.611 & 0.53 \\
        \bottomrule
	\end{tabular}
	\caption{English translation quality for all WMT Fr$\to$De ``$\to$English'' errors. Both translation directions are evaluated against English human references. *TER score is lower the better.}
	\label{tab:toen_quality}
\end{table}

Since none of our evaluated direction includes English as the target language, translating into English is never promoted and always trigger an off-target error.
We similarly sampled three ``$\to$English'' error cases from the WMT Fr$\to$De test set.
We also compare them against the real Fr$\to$En translations with the same model and French input.
This case study (in Table~\ref{tab:toen_cases}) hints that the ``$\to$English'' generations from WMT Fr$\to$De are generally \textit{similar} but slightly \textit{worse} ``English'' translations compared to Fr$\to$En.
To demonstrate the similarity, we plot the sentence BLEU distribution for all 172 ``$\to$English'' errors between Fr$\to$De and Fr$\to$En translations in Figure~\ref{fig:toen_bleu}.
It demonstrates a strong similarity between Fr$\to$De and Fr$\to$En translations, with an average sentence BLEU of 55.9.
Since the evaluation data of the WMT corpus is multi-way aligned, we can evaluate the $\to$English translation quality for both Fr$\to$De and Fr$\to$En against the English human references (in Table~\ref{tab:toen_quality}).
Results confirm our observation that the ``$\to$English'' errors are generally poorer English translations.

\subsection{Beam Search Process Analysis}

To understand how ``$\to$English'' and ``$\to$Source'' errors emerge during beam search and why both errors dramatically increase with larger beam sizes, we investigate the step-by-step beam search process with case studies.
Table~\ref{tab:beam_cases1} and \ref{tab:beam_cases2} illustrates one representative decoding example from the WMT Fr$\to$De test set with $b=5,20$ and French source ``Nous avons maintenant une excellente relation. »''.
For $b=20$, we only print the top-5 beams due to the space limit.
From this example, we have a few observations:
\begin{itemize}
    \item English candidates are live in the early steps (1-3) of $b=5$ but tend to be dropped in later time steps. Meanwhile for $b=20$, both English and French candidates are kept alive throughout the decoding process: even though they fall out of the top-5 beams at the 4th step, the off-target candidates quickly catch up and are ranked highest by the 7th step.
    \item Closely observing the winning English candidate of $b=20$, we notice it suffers a heavy penalty in the first step (log prob is -3.58), yet all following steps experience small penalties.
    \item The final English translation by $b=20$ is indeed a ``better'' candidate with greater probabilities (i.e. model score) compared to the final German translation by $b=5$, therefore, if this off-target candidate is retained throughout the process it will naturally win out against all valid on-target translations.
\end{itemize}

From the above observations, we can try to answer our previous research questions.
\begin{framed}
    \noindent {\bf RQ1}. How do ``$\to$English'' and ``$\to$Source'' errors emerge during decoding?
\end{framed}

We first observe that both ``$\to$English'' and ``$\to$Source'' candidates are easily accessible in the early steps of decoding.
Meanwhile, the models place a low probability on decoding the first source or English token, but relatively high transition probabilities for the remaining source or English tokens can result in off-target sentences scoring more highly than on-target.

\begin{framed}
    \noindent {\bf RQ2}. Why do both errors dramatically increase with larger beam sizes?
\end{framed}
With a larger beam size budget, it is more likely to retain off-target candidates in the earlier steps, when they receive heavy early step penalties. Yet since off-target candidates experience fewer penalties in the later steps, they tend to win out over on-target candidates in the long run.
We found it to be the general case that the off-target continuations receive a higher probability (less penalty) than the on-target ones, even though the first off-target token receives a heavy penalty by the model. We hypothesize that it is due to the \textit{recency bias} and poor calibration, yet it remains an interesting research question for future work.

\paragraph{Possible Solutions}
Off-target candidate gaining greater model score demonstrates that the model is poorly calibrated, especially for the later steps of autoregressive decoding.
Methods with additional training data~\cite{gu2019improved,zhang2020improving} or regularizations~\cite{yang2021improving} could alleviate this issue during training with a well-calibrated model.
In this work, with the knowledge of how off-target cases emerge during decoding, we attempt to fix this issue solely at the decoding time by improving on beam search algorithm even with a proven poorly calibrated model.

\section{Language-informed Beam Search (LiBS)}
The standard beam search process originating from the bilingual NMT model is target-language-agnostic and is found to produce an overwhelming number of off-target translations~\cite{zhang2020improving, yang2021improving}.
Yet the target language (i.e. the desired language for generation) is always known during decoding, thus it is straightforward to enforce the desired language to reduce the off-target rates without any additional training or data.
We thus propose Language-informed Beam Search (LiBS), a general decoding algorithm to inform the beam search process of the desired language during decoding.

\begin{algorithm2e*}[t!]
    \SetAlgoLined
    \DontPrintSemicolon
    \SetKwInOut{Input}{Input}
    \SetKwInOut{Output}{Output}
    \SetCommentSty{itshape}
    \SetKwComment{Comment}{$\triangleright$ }{}
    \Input{MNMT model $\theta$, LiD model $\gamma$, source sentence $\mathbf{x}$, target language $T$, beam size $b$, pre-select window size $w$}
    \Output{Finished candidate set $\mathbf{C} \gets \emptyset$}
    \Comment{Initialize each beam $i$ with BOS symbols and zero score}
    $\mathbf{B}_i \gets \{ \tuple{0.0,~\bos} \}$ \\
    \Repeat{$\mathbf{C}$ has $b$ finished candidates (i.e. $|\mathbf{C}|=b$)}{
        \Comment{Pre-select top-$w$ candidates from each beam $i$}
        $\mathbf{W}_i \gets \mathbf{top}_{w}\{ \tuple{s \cdot p_{\theta}(y~|~\mathbf{x}, \mathbf{y}),~\mathbf{y} \circ y} ~|~\tuple{s,~\mathbf{y}}\!\in\!\mathbf{B}_i,~y\!\in\!\mathcal{V} \}$ \\
        \Comment{Sort all candidates by the linearly combined NMT and LiD log probabilities}
        $\mathbf{W} \gets \mathbf{Sort} \{ \tuple{\log{s} + \alpha \log{p_{\gamma}(T,~\mathbf{y})},~s,~\mathbf{y}} ~|~ \tuple{s,~\mathbf{y}}\!\in\!\bigcup_{i=1}^{b}\mathbf{W}_i \}$ \label{alg:lid_rank_beam} \\
        \Comment{Store all finished ones from top-$b$ candidates into $\mathbf{C}$}
        $ \mathbf{C} \gets \{ \tuple{s,~\mathbf{y}}~|~ \tuple{s',s,~\mathbf{y}}\in \mathbf{top}_{b}\{ \mathbf{W} \},~\mathbf{y}_{|\mathbf{y}|} = \eos \} $ \\
        \Comment{Store the top-$b$ unfinished candidates into $\mathbf{B}$}
        $ \mathbf{B} \gets  \mathbf{top}_{b}\{ \tuple{s,~\mathbf{y}}~|~ \tuple{s',s,~\mathbf{y}}\in \mathbf{W},~\mathbf{y}_{|\mathbf{y}|} \neq \eos \} $ \\
    }
    \Comment{Rerank finished candidates by the linearly combined NMT and LiD log probabilities}
    $\mathbf{C} \gets \mathbf{Sort} \{ \tuple{\log{s} + \alpha \log{p_{\gamma}(T,~\mathbf{y})},~\mathbf{y}} ~|~ \tuple{s,~\mathbf{y}}\!\in\!\mathbf{C} \}$ \\
    \Return{C}
    \caption{Langugage-informed Beam Search}
    \label{alg:libs}
\end{algorithm2e*}

To inform the beam search process of the desired language, we borrow an off-the-shelf Language Identification (LiD) model to score the running beam search candidates with their probabilities in the correct language.
Since the candidates are normally ranked by the NMT model probabilities, we linearly combine the two log probabilities to ideally find the best candidate in the correct language.

The detailed algorithm is illustrated in Algorithm~\ref{alg:libs}.
For each step, we first pre-select top-$w$ candidates from each beam.
Then we sort all $b\cdot w$ active candidates by the linearly combined NMT and LiD log probabilities, where we tune the linear coefficient $\alpha$ on the dev set.
Same as the Fairseq~\cite{ott2019fairseq} implementation, we only store the finished ones within the top-$b$ candidates, meanwhile save the top-$b$ active candidates into the beam for the next step\footnote{We only store the NMT model score instead of the linearly combined one to avoid overcounting LiD scores.}.
The decoding process stops when we have found $b$ finished candidates, and at the end of the generation, we again rerank all finished candidates with the linearly combined log probabilities.

\paragraph{Design Choice and Speed Concern}
We only pre-select $b\cdot w$ candidates for the LiD scoring, instead of considering all possible continuations, simply because we could not afford to run LiD model on all $b \cdot |\mathcal{V}|$ candidates.

Even though in our experiments we only pre-select the top-2 continuations from each beam (i.e. $w=2$), the major slow down of the LiBS algorithm is still the un-BPE operation and LiD scoring on line~\ref{alg:lid_rank_beam} of Algorithm~\ref{alg:libs}.

To speed up the LiBS algorithm, we use the FastText LiD model since it is both fast and accurate (in our case on translation prefixes). With its help, LiBS is only 7.5 times slower than the Fairseq beam search decoding on a single CPU and 3.5 times slower with parallelized LiD scoring on 20 CPUs.

\section{Experiment Results}

To fully verify the performance of the LiBS algorithm, we compare LiBS against the baseline beam search decoding on both WMT and OPUS-100 datasets.

\subsection{WMT Results}

\paragraph{Tuning the Linear Coefficient $\alpha$}

\begin{table}[t]
	\centering
	\begin{tabular}{lcccc}
        \toprule
        \multirow{2}{*}{Model} & \multicolumn{2}{c}{De$\to$Fr} & \multicolumn{2}{c}{Cs$\to$De} \\
        \cmidrule{2-5}
         & \small BLEU & \small Off-Tgt & \small BLEU & \small Off-Tgt \\
        \midrule
        Baseline & 17.3 & 23.1\% & 15.4 & 12.3\% \\
        +LiBS, $\alpha=0.7$ & 20.6 & 2.0\% & 16.1 & 1.6\% \\
        +LiBS, $\alpha=0.8$ & 20.7 & 1.4\% & 16.2 & 1.6\% \\
        +LiBS, $\alpha=0.9$ & 20.7 & 1.1\% & 16.2 & 1.6\% \\
        +LiBS, $\alpha=1.0$ & 20.7 & 0.9\% & 16.2 & 1.4\% \\
        +LiBS, $\alpha=1.1$ & 20.7 & 0.9\% & 16.2 & 1.4\% \\
        +LiBS, $\alpha=1.2$ & 20.7 & 0.8\% & 16.2 & 1.3\% \\
        \bottomrule
	\end{tabular}
	\caption{Tuning the linear coefficient $\alpha$ on WMT De$\to$Fr and Cs$\to$De. }
	\label{tab:tune_coef}
\end{table}

We tune the linear coefficient $\alpha$ on the dev set. As shown in Table~\ref{tab:tune_coef}, any $\alpha$ value from 0.8 to 1.2 performs similarly well.
Because the linear coefficient $\alpha$ controls the weight of the LiD model score, as $\alpha$ increases, the off-target rate monotonically drops.
We use $\alpha=0.9$ for all the experiments on the WMT dataset.

\paragraph{Multilingual Beam Search Curse}

\begin{table*}[t]
	\centering
	\begin{tabular}{lc cccccc}
        \toprule
        \multicolumn{2}{c}{\multirow{2}{*}{Model}} & \multicolumn{3}{c}{De$\to$Fr} & \multicolumn{3}{c}{Cs$\to$De} \\
        \cmidrule{3-8}
         & & \small $b=5$ & \small $b=10$ & \small $b=20$  & \small $b=5$ & \small $b=10$ & \small $b=20$ \\
        \midrule
        \multirow{2}{*}{Baseline} & \small BLEU & 17.3 & 16.1 & 14.3 & 15.4 & 15.0 & 14.2 \\
        & \small Off-Tgt & 23.1\% & 31.7\% & 41.4\% & 12.3\% & 17.4\% &  22.0\%\\
        \midrule
        \multirow{2}{*}{+LiBS} & \small BLEU & 20.7 & 20.9 & 20.7 & 16.2 & 16.4 & 16.2 \\
        & \small Off-Tgt & 1.1\% & 1.2\% & 1.1\% & 1.6\% & 1.7\% & 1.5\%  \\
        \bottomrule
	\end{tabular}
	\caption{LiBS breaks the beam search curse on WMT De$\to$Fr and Cs$\to$De.}
	\label{tab:libs_curse}
\end{table*}

As illustrated before, the beam search curse exists in Multilingual NMT models predominantly due to the increasing off-target errors with larger beam sizes.
As shown in Table~\ref{tab:libs_curse}, LiBS successfully breaks the beam search curse by preventing off-target translations.

\begin{table*}[ht]
    \centering
    \setlength{\tabcolsep}{4pt}
    \resizebox{\textwidth}{!}{
    \begin{tabular}{lc cccccccccccc c}
    \toprule
    \multicolumn{2}{c}{\multirow{2}{*}{\textbf{Zero-Shot}}} & \multicolumn{2}{c}{Fr-De} & \multicolumn{2}{c}{De-Cs} & \multicolumn{2}{c}{Ro-De} & \multicolumn{2}{c}{Et-Fr}  & \multicolumn{2}{c}{Et-Ro} & \multicolumn{2}{c}{Gu-Tr} & \multirow{2}{*}{Average} \\
    & & $\gets$ & $\to$ & $\gets$ & $\to$ & $\gets$ & $\to$ & $\gets$ & $\to$ & $\gets$ & $\to$ & $\gets$ & $\to$ & \\
    \midrule
    \multirow{3}{*}{Baseline} & BLEU  & 17.3 & 11.7 & 15.4 & 13.9 & 17.2 & 16.1 & 10.6 & 13.5 & 11.9 & 14.1 & 0.9 & 2.0 & 12.05 \\
    & COMET & 0.72 & 0.74 & 0.71 & 0.76 & 0.60 & 0.70 & 0.72 & 0.74 & 0.53 & 0.74 & 0.75 & 0.48 & 0.68 \\
    & Off-Tgt & 23.1\% & 39.9\% & 12.3\% & 19.0\% & 1.6\% & 7.3\% & 22.5\% & 26.1\% & 10.8\% & 2.0\% & 73.7\% & 36.6\% & 22.91\%  \\
    \midrule
    \multirow{3}{*}{+LiBS} & BLEU  & 20.7 & 15.7 & 16.2 & 15.3 & 17.1 & 16.5 & 11.8 & 14.6 & 12.4 & 13.9 & 1.2 & 2.3 & 13.14 \\
    & COMET & 0.73 & 0.74 & 0.72 & 0.76 & 0.61 & 0.70 & 0.72 & 0.75 & 0.54 & 0.75 & 0.76 & 0.51 & 0.70 \\
    & Off-Tgt & 1.1\% & 6.5\% & 1.6\% & 3.8\% & 0.6\% & 0.6\% & 8.3\% & 4.6\% & 2.9\% & 0.3\% & 47.2\% & 15.0\% & 7.71\%  \\
    \bottomrule
    \end{tabular}}
    \caption{BLEU, COMET score and Off-Target rate of zero-shot translations on WMT dataset.}
    \label{tab:wmt_en_free}
\end{table*}

\paragraph{Zero-Shot Performance}
Table~\ref{tab:wmt_en_free} illustrates the full results of LiBS on the WMT dataset.
On average across all zero-shot directions, LiBS improves +1 BLEU score while reducing the off-target rates from 22.91\% to 7.71\%.
We notice that for many directions the off-target rate is barely around the error margin of the FastText LiD model, which is 0.81\% reported from ~\cite{yang2021improving}.
It hints that those translation directions do not suffer from off-target errors anymore, and the reported errors are largely due to the LiD model error.
Meanwhile, the MNMT model still suffers from a large number of off-target errors, especially on Gu$\to$Tr and Tr$\to$Gu translations, which we hypothesize is due to the extremely low resources for both languages (WMT contains 180K and 80K parallel data for Tr-En and Gu-En respectively.).

\subsection{OPUS-100 Results}

\begin{table*}[h!]
    \centering
    \setlength{\tabcolsep}{4pt}
    \resizebox{\textwidth}{!}{
    \begin{tabular}{lc cccccccccccc c}
    \toprule
    \multicolumn{2}{c}{\multirow{2}{*}{\textbf{Zero-Shot}}} & \multicolumn{2}{c}{De-Fr} & \multicolumn{2}{c}{Ru-Fr} & \multicolumn{2}{c}{Nl-De} & \multicolumn{2}{c}{Zh-Ru}  & \multicolumn{2}{c}{Zh-Ar} & \multicolumn{2}{c}{Nl-Ar} & \multirow{2}{*}{Average} \\
    & & $\gets$ & $\to$ & $\gets$ & $\to$ & $\gets$ & $\to$ & $\gets$ & $\to$ & $\gets$ & $\to$ & $\gets$ & $\to$ & \\
    \midrule
    \multirow{2}{*}{Baseline} & BLEU  & 3.3 & 3.0 & 5.4 & 4.0 & 5.9 & 5.2 & 5.7 & 11.8 & 3 & 11.6 & 1.2 & 3.2 & 5.28 \\
    & Off-Tgt & 95.2\% & 93.7\% & 68.9\% & 91.2\% & 88.4\% & 89.7\% & 37.0\% & 20.2\% & 89.9\% & 8.0\% & 93.0\% & 14.3\% & 65.79\%  \\
    \midrule
    \multirow{2}{*}{+LiBS} & BLEU  & 5.0 & 3.8 & 9.6 & 4.4 & 7.4 & 5.9 & 5.9 & 12.2 & 3.5 & 11.1 & 2.5 & 2.8 & 6.18  \\
    & Off-Tgt & 46.9\% & 49.5\% & 22.5\% & 41.6\% & 37.9\% & 40.0\% & 5.8\% & 1.1\% & 28.3\% & 0.7\% & 28.4\% & 1.4\% & 25.34\%  \\
    \bottomrule
    \end{tabular}}
    \caption{BLEU score and Off-Target rate of zero-shot translations on OPUS-100 dataset.}
    \label{tab:opus_en_free}
\end{table*}

\begin{figure}[t]\centering
    \includegraphics[width=0.49\textwidth]{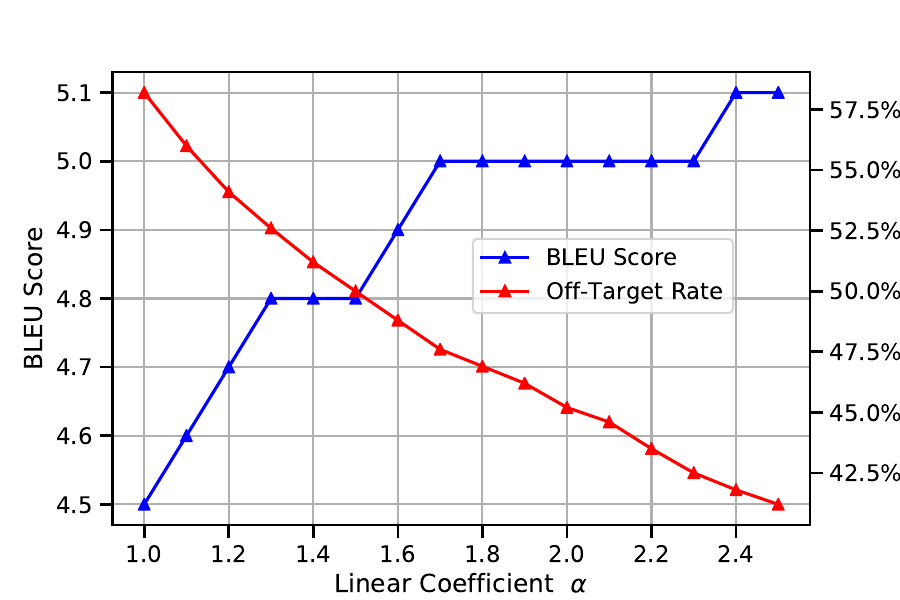}
    \caption{Translation performance (BLEU and off-target rates) with different $\alpha$ values on OPUS-100 Fr$\to$De test set.}
    \label{fig:alpha}
\end{figure}

To verify the effectiveness of our LiBS algorithm, we further compare it against the baseline beam search decoding on the large-scale OPUS-100 dataset, which includes a total of 100 languages.

Different from the WMT experiment, we tune and set $\alpha=1.8$ for all directions. This is due to the challenging nature of the OPUS-100 dataset that it performs very poorly on the zero-shot directions with a massive amount of off-target translations.
A higher $\alpha$ value for LiBS could effectively reduce the off-target rates and improve the translation performance.
For example, Figure~\ref{fig:alpha} plots the performance curve on the OPUS-100 Fr$\to$De test set with increasing $\alpha$ values.
It clearly shows a larger $\alpha$ value would consistently decrease the off-target rates and improve the overall performance (i.e. higher BLEU score)\footnote{The flat BLEU curve is due to the one decimal digit precision of sacreBLEU evaluation.}.

Zero-shot translation performance of LiBS on the OPUS-100 dataset is shown in Table~\ref{tab:opus_en_free}.
Across all directions, LiBS consistently improves an average of +0.9 BLEU and reduces the off-target rates from 65.79\% to 25.34\%.

Both WMT and OPUS-100 results clearly show our LiBS algorithm notably improves the zero-shot translation performance by significantly reducing the off-target translations.

\section{Related Work}
\paragraph{Off-Target Translation}
Off-target translation is a commonly observed failure mode in multilingual NMT models~\cite{arivazhagan2019missing}, and \citet{rios2020subword} has linked it to the predominance of English in the training data of multilingual models.
\citet{gu2019improved,zhang2019improving,yang2021improving} all observe it under different data settings and propose to mitigate it using   additional monolingual data or held-out oracle set.
Similarly to our work, \citet{sennrich2023mitigating} proposes to mitigate off-target errors with constrastive decoding, yet their approach usually hurts the translation quality, on average -1.1 BLEU on high resource languages.
Our work is the first to study off-target errors during decoding time, specifically how the off-target translations outscore on-target ones over time.
 
\paragraph{NMT Decoding}
Since beam search becomes the de-facto method for decoding NMT models~\cite{bahdanau2014neural}, studies has observed various flaws with it.
\citet{koehn2017six} observes beam search curse, where the translation quality usually degrades with increasing beam sizes.
\citet{yang2018breaking, stahlberg2019nmt} observe length bias, where the model heavily prefers shorter candidates.
To address those issues, extensive work has proposed sampling-based decoding algorithms, where the most popular one is Minimum Bayes Risk (MBR) decoding~\cite{eikema2020map}.
Yet, MBR decoding suffers severely from the quadratic complexity thus a slow inference speed.
Another line of research adopts external Language models to NMT beam search. Yet this external LM usually interferes with NMT's \textit{internal} LM (iLM).
However, with iLM neutralization, this approach still lags behind leveraging the additional monolingual data through back-translation~\cite{herold2023improving}.
Most similarly to our work, \citet{he2017decoding,ren2017deep} propose to incorporate a trained Value network during beam search decoding to improve the image-captioning task.
Our work instead attempts to mitigate off-target translation errors with a small off-the-shelf LiD model, while keeping the inference overhead to the linear scale (3.5x slow down).

\section{Conclusions}
Our work conducts a comprehensive off-target error analysis with strong multilingual NMT models, to answer the question of how off-target translation wins over time during decoding.
We additionally propose an empirical Language-informed Beam Search algorithm to mitigate off-target errors during decoding time and with linear-scale overhead.

\section{Limitations}
In this study, we utilize the widely adopted FastText LiD model, and the performance of LiBS may vary with the use of alternative LiD models.
As our method is a modified beam search algorithm, it is not directly applicable to recent Language Language Models~\cite{brown2020language}, which often do sampling during inference. 
Yet, we believe it will be particular interesting to adopt similar approach for LLM inference, as study shows LLMs are prune to hallucination~\cite{zhang2023siren}.

\bibliography{custom}

\begin{thebibliography}{27}
\expandafter\ifx\csname natexlab\endcsname\relax\def\natexlab#1{#1}\fi

\bibitem[{Arivazhagan et~al.(2019)Arivazhagan, Bapna, Firat, Aharoni, Johnson,
  and Macherey}]{arivazhagan2019missing}
Naveen Arivazhagan, Ankur Bapna, Orhan Firat, Roee Aharoni, Melvin Johnson, and
  Wolfgang Macherey. 2019.
\newblock The missing ingredient in zero-shot neural machine translation.
\newblock \emph{arXiv preprint arXiv:1903.07091}.

\bibitem[{Bahdanau et~al.(2014)Bahdanau, Cho, and Bengio}]{bahdanau2014neural}
Dzmitry Bahdanau, Kyunghyun Cho, and Yoshua Bengio. 2014.
\newblock Neural machine translation by jointly learning to align and
  translate.
\newblock \emph{arXiv preprint arXiv:1409.0473}.

\bibitem[{Brown et~al.(2020)Brown, Mann, Ryder, Subbiah, Kaplan, Dhariwal,
  Neelakantan, Shyam, Sastry, Askell et~al.}]{brown2020language}
Tom Brown, Benjamin Mann, Nick Ryder, Melanie Subbiah, Jared~D Kaplan, Prafulla
  Dhariwal, Arvind Neelakantan, Pranav Shyam, Girish Sastry, Amanda Askell,
  et~al. 2020.
\newblock Language models are few-shot learners.
\newblock \emph{Advances in neural information processing systems},
  33:1877--1901.

\bibitem[{Eikema and Aziz(2020)}]{eikema2020map}
Bryan Eikema and Wilker Aziz. 2020.
\newblock Is map decoding all you need? the inadequacy of the mode in neural
  machine translation.
\newblock \emph{arXiv preprint arXiv:2005.10283}.

\bibitem[{Gu et~al.(2019)Gu, Wang, Cho, and Li}]{gu2019improved}
Jiatao Gu, Yong Wang, Kyunghyun Cho, and Victor~OK Li. 2019.
\newblock Improved zero-shot neural machine translation via ignoring spurious
  correlations.
\newblock \emph{arXiv preprint arXiv:1906.01181}.

\bibitem[{He et~al.(2017)He, Lu, Xia, Qin, Wang, and Liu}]{he2017decoding}
Di~He, Hanqing Lu, Yingce Xia, Tao Qin, Liwei Wang, and Tie-Yan Liu. 2017.
\newblock Decoding with value networks for neural machine translation.
\newblock In \emph{Advances in Neural Information Processing Systems}, pages
  178--187.

\bibitem[{Herold et~al.(2023)Herold, Gao, Zeineldeen, and
  Ney}]{herold2023improving}
Christian Herold, Yingbo Gao, Mohammad Zeineldeen, and Hermann Ney. 2023.
\newblock Improving language model integration for neural machine translation.
\newblock \emph{arXiv preprint arXiv:2306.05077}.

\bibitem[{Johnson et~al.(2017)Johnson, Schuster, Le, Krikun, Wu, Chen, Thorat,
  Vi{\'e}gas, Wattenberg, Corrado et~al.}]{johnson2017google}
Melvin Johnson, Mike Schuster, Quoc~V Le, Maxim Krikun, Yonghui Wu, Zhifeng
  Chen, Nikhil Thorat, Fernanda Vi{\'e}gas, Martin Wattenberg, Greg Corrado,
  et~al. 2017.
\newblock Google’s multilingual neural machine translation system: Enabling
  zero-shot translation.
\newblock \emph{Transactions of the Association for Computational Linguistics},
  5:339--351.

\bibitem[{Joulin et~al.(2016)Joulin, Grave, Bojanowski, Douze, J{\'e}gou, and
  Mikolov}]{joulin2016fasttext}
Armand Joulin, Edouard Grave, Piotr Bojanowski, Matthijs Douze, H{\'e}rve
  J{\'e}gou, and Tomas Mikolov. 2016.
\newblock Fasttext.zip: Compressing text classification models.
\newblock \emph{arXiv preprint arXiv:1612.03651}.

\bibitem[{Kingma and Ba(2015)}]{kingma2014adam}
Diederik~P Kingma and Jimmy Ba. 2015.
\newblock Adam: A method for stochastic optimization.
\newblock In \emph{ICLR}.

\bibitem[{Koehn and Knowles(2017)}]{koehn2017six}
Philipp Koehn and Rebecca Knowles. 2017.
\newblock Six challenges for neural machine translation.
\newblock \emph{arXiv preprint arXiv:1706.03872}.

\bibitem[{Kudo and Richardson(2018)}]{kudo2018sentencepiece}
Taku Kudo and John Richardson. 2018.
\newblock Sentencepiece: A simple and language independent subword tokenizer
  and detokenizer for neural text processing.
\newblock \emph{arXiv preprint arXiv:1808.06226}.

\bibitem[{Ott et~al.(2018)Ott, Auli, Grangier, and Ranzato}]{ott2018analyzing}
Myle Ott, Michael Auli, David Grangier, and Marc’Aurelio Ranzato. 2018.
\newblock Analyzing uncertainty in neural machine translation.
\newblock In \emph{International Conference on Machine Learning}, pages
  3956--3965. PMLR.

\bibitem[{Ott et~al.(2019)Ott, Edunov, Baevski, Fan, Gross, Ng, Grangier, and
  Auli}]{ott2019fairseq}
Myle Ott, Sergey Edunov, Alexei Baevski, Angela Fan, Sam Gross, Nathan Ng,
  David Grangier, and Michael Auli. 2019.
\newblock fairseq: A fast, extensible toolkit for sequence modeling.
\newblock In \emph{NAACL-HLT}.

\bibitem[{Post(2018)}]{post2018call}
Matt Post. 2018.
\newblock A call for clarity in reporting bleu scores.
\newblock \emph{arXiv preprint arXiv:1804.08771}.

\bibitem[{Rei et~al.(2020)Rei, Stewart, Farinha, and Lavie}]{rei2020comet}
Ricardo Rei, Craig Stewart, Ana~C Farinha, and Alon Lavie. 2020.
\newblock Comet: A neural framework for mt evaluation.
\newblock \emph{arXiv preprint arXiv:2009.09025}.

\bibitem[{Ren et~al.(2017)Ren, Wang, Zhang, Lv, and Li}]{ren2017deep}
Zhou Ren, Xiaoyu Wang, Ning Zhang, Xutao Lv, and Li-Jia Li. 2017.
\newblock Deep reinforcement learning-based image captioning with embedding
  reward.
\newblock In \emph{Proceedings of the IEEE conference on computer vision and
  pattern recognition}, pages 290--298.

\bibitem[{Rios et~al.(2020)Rios, M{\"u}ller, and Sennrich}]{rios2020subword}
Annette Rios, Mathias M{\"u}ller, and Rico Sennrich. 2020.
\newblock Subword segmentation and a single bridge language affect zero-shot
  neural machine translation.
\newblock \emph{arXiv preprint arXiv:2011.01703}.

\bibitem[{Sennrich et~al.(2023)Sennrich, Vamvas, and
  Mohammadshahi}]{sennrich2023mitigating}
Rico Sennrich, Jannis Vamvas, and Alireza Mohammadshahi. 2023.
\newblock Mitigating hallucinations and off-target machine translation with
  source-contrastive and language-contrastive decoding.
\newblock \emph{arXiv preprint arXiv:2309.07098}.

\bibitem[{Stahlberg and Byrne(2019)}]{stahlberg2019nmt}
Felix Stahlberg and Bill Byrne. 2019.
\newblock On nmt search errors and model errors: Cat got your tongue?
\newblock \emph{arXiv preprint arXiv:1908.10090}.

\bibitem[{Vaswani et~al.(2017)Vaswani, Shazeer, Parmar, Uszkoreit, Jones,
  Gomez, Kaiser, and Polosukhin}]{vaswani2017attention}
Ashish Vaswani, Noam Shazeer, Niki Parmar, Jakob Uszkoreit, Llion Jones,
  Aidan~N Gomez, Lukasz Kaiser, and Illia Polosukhin. 2017.
\newblock Attention is all you need.
\newblock \emph{arXiv preprint arXiv:1706.03762}.

\bibitem[{Wang et~al.(2020)Wang, Zhai, and Awadalla}]{wang2020multi}
Yiren Wang, ChengXiang Zhai, and Hany~Hassan Awadalla. 2020.
\newblock Multi-task learning for multilingual neural machine translation.
\newblock \emph{arXiv preprint arXiv:2010.02523}.

\bibitem[{Yang et~al.(2021)Yang, Eriguchi, Muzio, Tadepalli, Lee, and
  Hassan}]{yang2021improving}
Yilin Yang, Akiko Eriguchi, Alexandre Muzio, Prasad Tadepalli, Stefan Lee, and
  Hany Hassan. 2021.
\newblock Improving multilingual translation by representation and gradient
  regularization.
\newblock \emph{arXiv preprint arXiv:2109.04778}.

\bibitem[{Yang et~al.(2018)Yang, Huang, and Ma}]{yang2018breaking}
Yilin Yang, Liang Huang, and Mingbo Ma. 2018.
\newblock Breaking the beam search curse: A study of (re-) scoring methods and
  stopping criteria for neural machine translation.
\newblock \emph{arXiv preprint arXiv:1808.09582}.

\bibitem[{Zhang et~al.(2019)Zhang, Titov, and Sennrich}]{zhang2019improving}
Biao Zhang, Ivan Titov, and Rico Sennrich. 2019.
\newblock Improving deep transformer with depth-scaled initialization and
  merged attention.
\newblock \emph{arXiv preprint arXiv:1908.11365}.

\bibitem[{Zhang et~al.(2020)Zhang, Williams, Titov, and
  Sennrich}]{zhang2020improving}
Biao Zhang, Philip Williams, Ivan Titov, and Rico Sennrich. 2020.
\newblock Improving massively multilingual neural machine translation and
  zero-shot translation.
\newblock \emph{arXiv preprint arXiv:2004.11867}.

\bibitem[{Zhang et~al.(2023)Zhang, Li, Cui, Cai, Liu, Fu, Huang, Zhao, Zhang,
  Chen et~al.}]{zhang2023siren}
Yue Zhang, Yafu Li, Leyang Cui, Deng Cai, Lemao Liu, Tingchen Fu, Xinting
  Huang, Enbo Zhao, Yu~Zhang, Yulong Chen, et~al. 2023.
\newblock Siren's song in the ai ocean: a survey on hallucination in large
  language models.
\newblock \emph{arXiv preprint arXiv:2309.01219}.

\end{thebibliography}
\bibliographystyle{acl_natbib}

\appendix

\section{Additional Tables}
\begin{table*}[ht]
	\centering
	\begin{tabularx}{\textwidth}{LYLKYLK}
        \toprule
	    & \multicolumn{3}{c}{$b=5$} & \multicolumn{3}{c}{$b=20$} \\
        \cmidrule(lr){2-4}\cmidrule(lr){5-7}
        Step & Beam & LiD & LogProb & Beam & LiD & LogProb \\
        \midrule
        \multirow{5}{*}{1}&\_Wir&De&-1.10 & \_Wir&De&-1.10\\
        & \_"&En&-2.93 &\_"&En&-2.93\\
        &\_Jetzt&De&-3.02&\_Jetzt&De&-3.02\\
        &\textcolor{red}{\_We}&En&-3.58&\textcolor{red}{\_We}&En&-3.58\\
        &\_„&Ro&-3.61&\_„&Ro&-3.61\\
        \midrule
        \multirow{5}{*}{2}&\_Wir \_haben&De&-1.44&\_Wir \_haben&De&-1.44\\
        &\_Jetzt \_haben&De&-3.48&\_Jetzt \_haben&De&-3.48\\
        &\textcolor{red}{\_We \_now}&En&-4.10&\textcolor{red}{\_We \_now}&En&-4.10\\
        &\_Wir \_verfügen&De&-4.29&\textcolor{blue}{\_Nous \_avons}&Fr&-4.13\\
        &\textcolor{red}{\_We \_have}&En&-4.83&\_Wir \_verfügen&De&-4.29\\
        \midrule
        \multirow{5}{*}{3}&\_Wir \_haben \_jetzt&De&-2.22&\_Wir \_haben \_jetzt&De&-2.22\\
        &\_Wir \_haben \_nun&De&-2.82&\_Wir \_haben \_nun&De&-2.82\\
        &\_Jetzt \_haben \_wir&De&-3.60&\_Jetzt \_haben \_wir&De&-3.60\\
        &\textcolor{red}{\_We \_now \_have}&En&-4.20&\textcolor{red}{\_We \_now \_have}&En&-4.20\\
        &\_Wir \_haben \_eine&De&-4.83&\textcolor{blue}{\_Nous \_avons \_maintenant}&Fr&-4.36\\
        \midrule
        \multirow{5}{*}{4}&\_Wir \_haben \_jetzt \_eine&De&-2.65&\_Wir \_haben \_jetzt \_eine&De&-2.65\\
        &\_Wir \_haben \_nun \_eine&De&-3.24&\_Wir \_haben \_nun \_eine&De&-3.24\\
        &\_Wir \_haben \_jetzt \_ein&De&-3.93&\_Wir \_haben \_jetzt \_ein&De&-3.93\\
        &\_Jetzt \_haben \_wir \_eine&De&-4.12&\_Jetzt \_haben \_wir \_eine&De&-4.12\\
        &\_Wir \_haben \_nun \_ein&De&-4.58&\_Wir \_haben \_nun \_ein&De&-4.58\\
        \midrule
        \multirow{5}{*}{5}&\_Wir \_haben \_jetzt \_eine \_ausgezeichnete&De&-3.66&\_Wir \_haben \_jetzt \_eine \_ausgezeichnete&De&-3.66\\
        &\_Wir \_haben \_nun \_eine \_ausgezeichnete&De&-4.23&\_Wir \_haben \_nun \_eine \_ausgezeichnete&De&-4.23\\
        &\_Wir \_haben \_jetzt \_eine \_hervorragende&De&-4.25&\_Wir \_haben \_jetzt \_eine \_hervorragende&De&-4.25\\
        &\_Wir \_haben \_nun \_eine \_hervorragende&De&-4.81&\textcolor{red}{\_We \_now \_have \_an \_excellent}&En&-4.80\\
        &\_Wir \_haben \_jetzt \_eine \_exzellente&De&-4.95&\_Wir \_haben \_nun \_eine \_hervorragende&De&-4.81\\
        \bottomrule
	\end{tabularx}
	\caption{Beam Search case study for $b=5$ and $b=20$ on one example from WMT Fr$\to$De test set. English candidates (``$\to$English'' errors) are colored in red, while French candidates (``$\to$Source'' errors) are colored in blue.}
	\label{tab:beam_cases1}
\end{table*}

\begin{table*}[ht]
	\centering
    \small
	\begin{tabularx}{\textwidth}{LYLKYLK}
        \toprule
	    & \multicolumn{3}{c}{$b=5$} & \multicolumn{3}{c}{$b=20$} \\
        \cmidrule(lr){2-4}\cmidrule(lr){5-7}
        Step & Beam & LiD & LogProb & Beam & LiD & LogProb \\
        \multirow{5}{*}{6}&\_Wir \_haben \_jetzt \_eine \_ausgezeichnete \_Beziehung&De&-3.82&\_Wir \_haben \_jetzt \_eine \_ausgezeichnete \_Beziehung&De&-3.82\\
        &\_Wir \_haben \_nun \_eine \_ausgezeichnete \_Beziehung&De&-4.39&\_Wir \_haben \_nun \_eine \_ausgezeichnete \_Beziehung&De&-4.39\\
        &\_Wir \_haben \_jetzt \_eine \_hervorragende \_Beziehung&De&-4.42&\_Wir \_haben \_jetzt \_eine \_hervorragende \_Beziehung&De&-4.42\\
        &\_Wir \_haben \_nun \_eine \_hervorragende \_Beziehung&De&-4.98&\textcolor{red}{\_We \_now \_have \_an \_excellent \_relationship}&En&-4.89\\
        &\_Wir \_haben \_jetzt \_eine \_exzellente \_Beziehung&De&-5.12&\_Wir \_haben \_nun \_eine \_hervorragende \_Beziehung&De&-4.98\\
        \midrule
        \multirow{5}{*}{7}&\textbf{\_Wir \_haben \_jetzt \_eine \_ausgezeichnete \_Beziehung .“}&De&-5.73&\textcolor{blue}{\_Nous \_avons \_maintenant \_une \_excellent e \_relation}&Fr&-5.53\\
        &\_Wir \_haben \_jetzt \_eine \_ausgezeichnete \_Beziehung .&De&-5.89&\_Wir \_haben \_jetzt \_ein \_ausgezeichnete s \_Verhältnis&De&-5.54\\
        &\_Wir \_haben \_jetzt \_eine \_ausgezeichnete \_Beziehung ."&De&-5.95&\textbf{\textcolor{red}{\_We \_now \_have \_an \_excellent \_relationship ."}}&En&-5.70\\
        &\_Wir \_haben \_jetzt \_eine \_hervorragende \_Beziehung .“&De&-6.28&\_Wir \_haben \_jetzt \_eine \_ausgezeichnete \_Beziehung .“&De&-5.73\\
        &\_Wir \_haben \_nun \_eine \_ausgezeichnete \_Beziehung .“&De&-6.31&\_Wir \_haben \_jetzt \_ein \_hervorragende s \_Verhältnis&De&-5.82\\
        \bottomrule
	\end{tabularx}
	\caption{Beam Search case study for $b=5$ and $b=20$ on one example from WMT Fr$\to$De test set. English candidates (``$\to$English'' errors) are colored in red, while French candidates (``$\to$Source'' errors) are colored in blue. Final translations (at step 7) are in bold, where $b=5$ generates a German translation, and $b=20$ generates an off-target English translation at the 7th step.}
	\label{tab:beam_cases2}
\end{table*}

\end{document}